%% file: main.tex
\documentclass[10pt,twocolumn,letterpaper]{article}

\usepackage[pagenumbers]{cvpr} 

\input{preamble}
\usepackage{tikz}
\usepackage{svg} 
\usetikzlibrary{arrows.meta, positioning, fit, calc}
\definecolor{cvprblue}{rgb}{0.21,0.49,0.74}
\usepackage[pagebackref,breaklinks,colorlinks,allcolors=cvprblue]{hyperref}
\input{macros}
\def\paperID{} 
\def\confName{CVPR}
\def\confYear{2026}

\title{\modelname: Camera-Controllable Video Generation via \\ Consistent and Extensible Tokenization}

\author{
Zelin Zhao$^{1}$ \quad
Xinyu Gong$^{2}$ \quad
Bangya Liu$^{3}$ \quad
Ziyang Song$^{4}$ \quad
Jun Zhang$^{2}$ \quad \\
Suhui Wu$^{2}$ \quad
Yongxin Chen$^{1}$ \quad
Hao Zhang$^{2}$ \\
[4pt]
$^{1}$Georgia Institute of Technology \quad
$^{2}$ByteDance \quad \\
$^{3}$University of Wisconsin--Madison \quad
$^{4}$The Hong Kong Polytechnic University
}

\begin{document}
\twocolumn[{%
\renewcommand\twocolumn[1][]{#1}%
\maketitle
\begin{center}
    \centering
    \vspace{-18pt}
    \captionsetup{type=figure}
\includegraphics[width=1.0\textwidth]{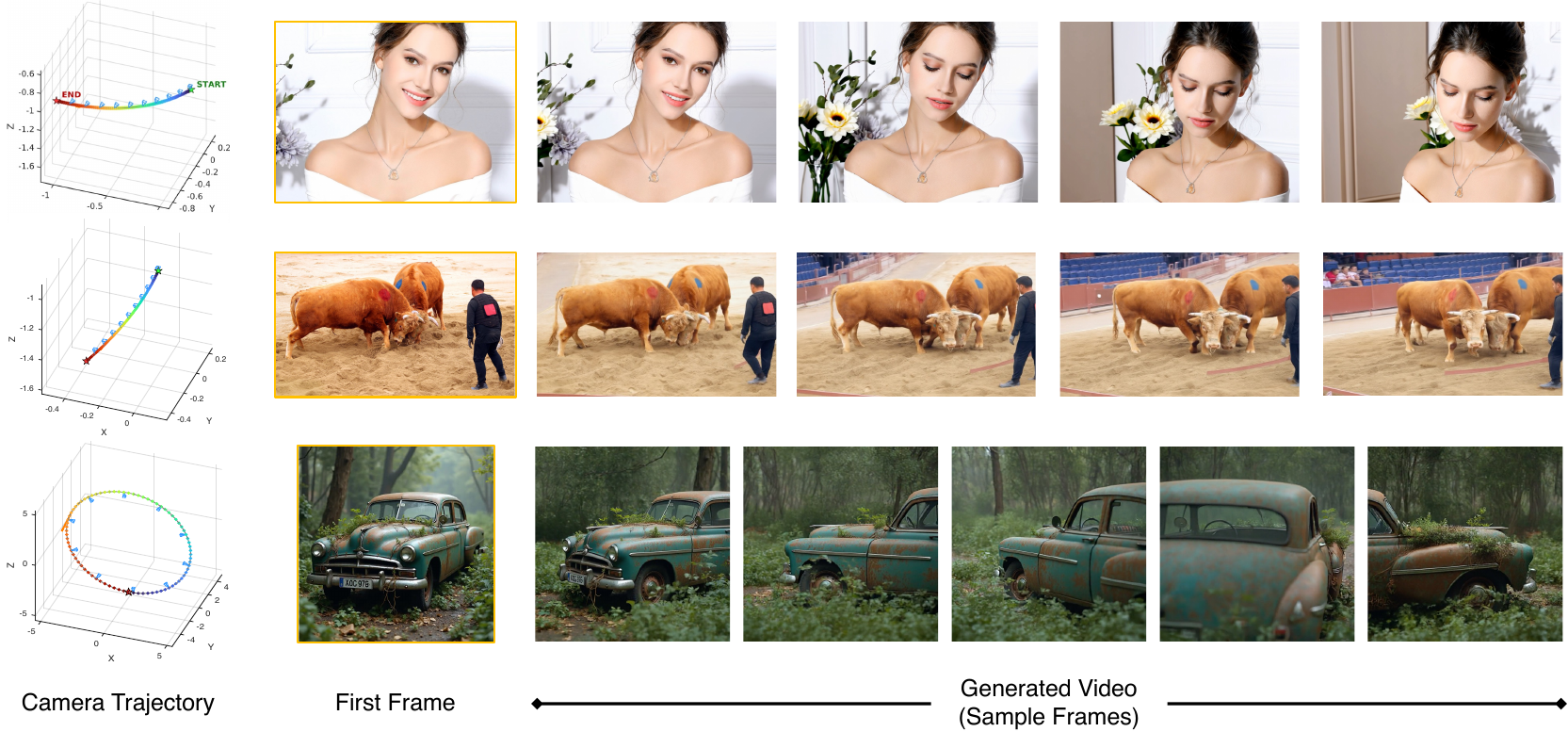}
    \vspace{-15pt}
    \captionof{figure}{
\textbf{Camera-controlled Video Generation.} 
Our framework synthesizes \textbf{dynamic}, \textbf{geometry-consistent} scenes across diverse domains, including human face, animal motion, and natural environments. It allows accurate camera control across wide viewpoints, with full 360° orbit motion shown in the last example. Please refer to the project page for videos, code and pre-trained models.
}
\label{fig-teaser}
\end{center}%
}]
\input{sec/0_abstract}    
\input{sec/1_intro}
\input{sec/2_related_works}
\input{sec/3_finalcopy}

\newpage
\clearpage
{
    \small
    \bibliographystyle{ieeenat_fullname}
    \bibliography{main}
}

\input{sec/X_suppl}

\end{document}

%% file: macros.tex
\usepackage{xspace}
\usepackage{amsmath}
\usepackage{amssymb}
\usepackage{mathtools}
\usepackage{amsthm}
\usepackage[table]{xcolor}
\usepackage{algorithm}
\usepackage{algorithmic}
\usepackage[capitalize,noabbrev]{cleveref}
\Crefname{ALC@unique}{Line}{Lines}
\theoremstyle{plain}

\theoremstyle{definition}

\theoremstyle{remark}

\definecolor{blue}{HTML}{49a0d4}
\definecolor{red}{HTML}{cc1100}
\definecolor{orange}{HTML}{cc7700}
\definecolor{gray}{HTML}{efefef}
\definecolor{darkgreen}{HTML}{228B22}
\definecolor{darkgray}{HTML}{757575}

\newcommand{\modelname}{\textsc{CETCam}\xspace}
\newcommand{\datasetname}{HoIHQ\xspace}
\newcommand{\vggt}{\textsc{VGGT}\xspace}
\newcommand{\unic}{\textsc{Uni3C}\xspace}
\newcommand{\vace}{\textsc{VACE}\xspace}

\definecolor{grey}{rgb}{0.6,0.6,0.6}
\definecolor{lightgray}{rgb}{0.97,.99,0.99}

%% file: sec/0_abstract.tex
\begin{abstract}
Achieving precise camera control in video generation remains challenging, as it often relies on camera pose annotations that are difficult to scale to large, dynamic datasets and are frequently inconsistent with depth estimation, leading to train–test discrepancies. We introduce \modelname, a camera-controllable video generation framework that eliminates the need for camera annotations through consistent and extensible tokenization. \modelname incorporates recent advances in geometry foundation models (e.g., VGGT) to estimate depth and camera parameters, converting them into unified, geometry-aware tokens that are seamlessly integrated into a pretrained video diffusion backbone via lightweight context blocks. Trained in two progressive phases, \modelname learns robust camera controllability from diverse raw videos and refines fine-grained visual quality on curated high-fidelity data. Extensive experiments across multiple benchmarks demonstrate state-of-the-art geometric consistency, temporal stability, and visual realism. Moreover, \modelname exhibits strong adaptability to additional control modalities (e.g., inpainting or layout), highlighting its flexibility beyond camera control. The project page is available at \url{https://sjtuytc.github.io/CETCam_project_page.github.io/}.
\end{abstract}

%% file: sec/1_intro.tex
\section{Introduction}
\label{sec:intro}
Recent advances in video diffusion models (VDMs) have greatly improved visual quality and temporal coherence across diverse domains~\cite{wan2025,yang2025cogvideox}. However, precise camera control in dynamic scenes remains a major challenge. Existing large-scale video foundation models, such as Wan~\cite{wan2025} and CogVideoX~\cite{yang2025cogvideox}, often lack explicit 3D understanding and struggle to manipulate viewpoints reliably. As a result, generating videos with consistent multi-view geometry or user-specified camera motion remains highly non-trivial—let alone combining camera control with other conditioning tasks such as object manipulation or inpainting.

Earlier attempts to address this challenge can be grouped into three broad categories.
Training-free approaches~\cite{camtrol,latentReframe} manipulate pretrained diffusion models via latent-space optimization or editing.
While simple to deploy, these methods are typically unstable and fail to generalize to unseen scenes or complex camera trajectories~\cite{cao2025uni3c}.
Novel-view synthesis (NVS) methods~\cite{yu2024viewcrafter,seo2024genwarp,multidiff,CamCtrl3D} introduce explicit 3D constraints by reconstructing and rendering scene geometry.
However, they are largely limited to static environments and cannot effectively model dynamic motion or long-range temporal consistency. More recently, several camera-controllable diffusion frameworks such as ReCamMaster~\cite{recammaster} and Gen3C~\cite{ren2025gen3c} have extended diffusion-based video generation with explicit camera conditioning, often by constructing or rendering intermediate 3D representations. While these approaches demonstrate improved viewpoint control, they typically train on annotated camera poses or separately estimated depth supervision. Moreover, their designs are specialized for camera motion only—making them difficult to extend to other conditioning modalities such as sketch or layout. Furthermore, the concurrent work Uni3C~\cite{cao2025uni3c} explores a similar idea of integrating camera control into video diffusion models but often suffers from local distortions and geometric misalignment due to inaccurate 3D consistency between estimated depth and pose signals.

To overcome these limitations, we propose \textbf{C}onsistent and \textbf{E}xtensible \textbf{T}okenization (\textbf{\modelname}) for controllable camera-driven video generation. 
Our model builds upon recent advances of the Visual Geometry Grounded Transformer (VGGT)~\cite{vggt}, jointly predicting camera poses and depth maps, followed by point-based reprojection that generates renderings and masks that encode view-dependent geometry. 
These signals are then embedded into compact, geometry-aware camera tokens, enabling explicit yet annotation-free camera conditioning within the diffusion process. 
To integrate with large-scale video diffusion backbones, we introduce \modelname context blocks that serve as plug-in modules to the frozen base video diffusion model, allowing seamless conditioning without retraining the base model. 
As shown in \Cref{fig-teaser}, our framework enables precise camera control across diverse scenes while preserving geometric consistency and visual fidelity.

Beyond camera control, our framework is inherently extensible. Inspired by the \vace architecture~\cite{vace}, \modelname integrates seamlessly with additional control modalities, supporting unified compositional generation under multiple types of conditioning. 
This design allows \modelname\ to go beyond camera control and serve as a general interface for combining geometric and semantic guidance in controllable video generation.

Extensive experiments demonstrate that \modelname significantly outperforms prior camera-controllable video generation methods across multiple benchmarks. 
Compared with Uni3C~\cite{cao2025uni3c}, our approach achieves overall VBench~\cite{vbench} score improvements of +3.09\%, +2.70\%, and +2.88\% on the Uni3C-OOD-Challenging~\cite{cao2025uni3c}, CameraBench~\cite{camerabench}, and our newly collected \datasetname benchmark, respectively. 
The \datasetname dataset is a high-fidelity human--object interaction benchmark specifically curated for evaluating camera-controllable video generation in dynamic and realistic settings. 
Human evaluation results further confirm the superiority of our method, with perceptual scores of 90.3, 91.5, and 93.4 on the respective benchmarks---an average gain exceeding 6\% over the previous best. 
Additionally, \modelname achieves more accurate and stable camera trajectory following, reducing pose errors by over 15\% across all benchmarks. 
Moreover, the framework preserves strong camera controllability and high generative quality while remaining extensible to additional control modalities.

\noindent Our \textbf{main contributions} are summarized as follows:
\begin{enumerate}
    \item We propose the~\modelname camera tokenizer for controllable viewpoint video generation, where it jointly predicts depth maps and camera parameters, and performs point-based reprojection to generate renderings and masks that encode view-dependent geometry. These signals are further embedded into camera tokens, eliminating the need for explicit camera labels and enabling robust training across diverse real-world videos.
   \item We introduce an extensible controlled video generation framework that integrates the proposed camera tokens with additional conditioning modalities. The design enables seamless compositional control—combining camera motion with inputs such as inpainting or grayscale cues—while preserving the expressive and generative capacity of the base diffusion model. This establishes a scalable, token-based conditioning interface applicable across diverse control types.
   \item Through extensive experiments across multiple benchmarks, we demonstrate that our method achieves accurate camera control, maintaining strong temporal and geometric consistency across diverse scenes. It also preserves high visual fidelity compared to recent baselines and further validates the framework’s extensibility when integrated with other controllable generation modalities.
\end{enumerate}

%% file: sec/2_related_works.tex
\section{Related Works}
\label{sec-related-works}
\paragraph{Video Generation with Camera Control.} 
Camera-controllable video generation has recently moved from implicit viewpoint conditioning to explicit 3D-aware control. Training-free methods are often limited in quality or extensibility~\cite{camtrol, latentReframe}. Most video novel view synthesis (NVS) methods are limited to static scenes and cannot model dynamic motions~\cite{yu2024viewcrafter, CamCtrl3D, seo2024genwarp,multidiff, seva}. Gen3C~\cite{ren2025gen3c} introduces the idea of constructing and rendering a 3D cache as a conditioning signal, outperforming approaches that directly inject camera poses~\cite{cameractrl}. Similar 3D rendering–based mechanisms are adopted in recent works~\cite{cao2025uni3c, Ex4D, epic, bahmani2025ac3d, cameractrl}, where depth or point-based re-projections are used to inject camera pose into the diffusion backbone. EPiC~\cite{epic}, though architecturally distinct from ours, is trained on a very small dataset and built upon a weaker base model (CogVideoX-5B~\cite{yang2025cogvideox}), leading to inferior visual fidelity. The closest concurrent work, \unic~\cite{cao2025uni3c}, depends on camera annotations and adopts mismatched depth estimators during training (SFM~\cite{sfm}) and testing (Depth Pro~\cite{depthpro}). Our method differs by learning directly from raw dynamic videos without requiring annotation, achieving consistent geometric grounding, better visual quality and higher extensibility.
\begin{figure*}[t!]
\begin{center}
\includegraphics[width=1.0\linewidth]{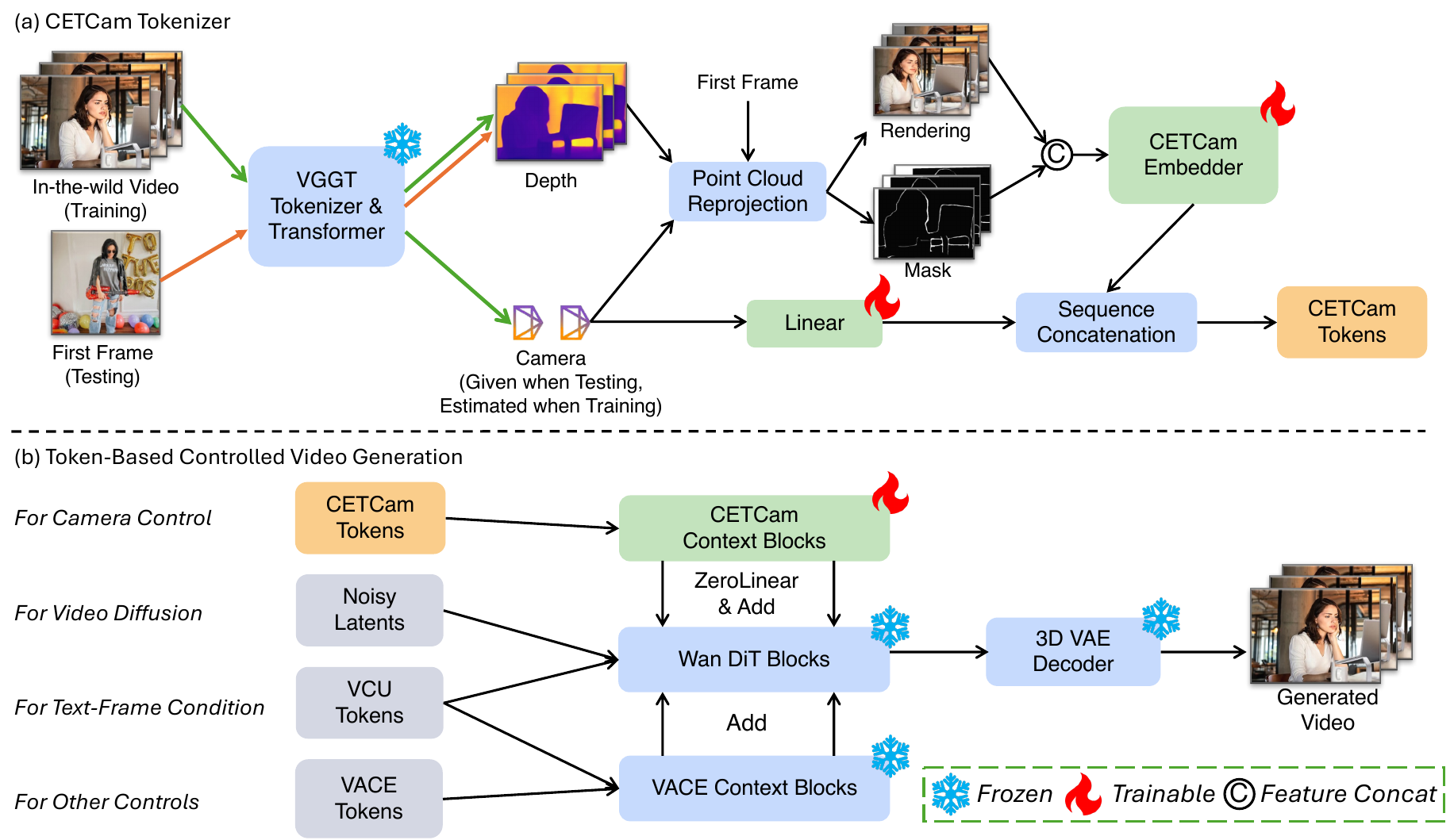}
\end{center}
\vspace{-10pt}
\caption{
\textbf{Overview of \modelname I2V generation framework.}
(a) \textit{\modelname Tokenizer.}
Given an in-the-wild training video or a test-time frame input, the frames are processed by \vggt\cite{vggt} to predict the depth maps. In training, camera poses are also estimated. Predicted depths and camera poses are used for point cloud reprojection to generate renderings of the first frame and corresponding masks. Renderings, masks, and camera poses are embedded and fused to produce \modelname tokens. 
(b) \textit{Token-Based Controlled Video Generation.} We leverage various tokens with rich and diverse functions via different tokens, including \modelname tokens, noisy latents, VCU tokens, and \vace tokens~\cite{vace}. \modelname tokens are consumed in learnable \modelname context blocks, which were connected to pre-trained Wan DiT blocks~\cite{wan2025} with zero linear and add functions~\cite{cao2025uni3c}. Other tokens are further processed by \vace context blocks and Wan DiT blocks. Finally, the output tokens are decoded through a 3D VAE to generate a video~\cite{wan2025}. More details can be found in~\Cref{sec-tokenizer,sec-controlledgen}.}
\label{fig-model-ours}
\vspace{-6pt}
\end{figure*}
\vspace{-3mm}
\paragraph{Controllable Video Synthesis and Editing.}
Beyond viewpoint control, a parallel line of work focuses on controllable video diffusion models that accept diverse user controls. VideoCrafter2~\cite{chen2024videocrafter2} and VideoComposer~\cite{videocomposer} extends text-to-video diffusion models with various motion control techniques, while ControlVideo~\cite{zhang2024controlvideo} adapts image-control modules to preserve temporal consistency. \vace~\cite{vace} is a unified model that extends ControlNet-style conditioning to videos, enabling versatile and composable video generation and editing tasks within a single framework. MotionDirector~\cite{zhao2024motiondirector} does not address the more challenging problem of controlling camera angle changes, which remains nontrivial. FullDiT~\cite{fulldit}, which is a recent concurrent work without available pre-trained models or code, conducts various controls including camera control.
\vspace{-4mm}
\paragraph{Geometry-grounded Models.} Recent works focus on learning geometry from visual data through feed-forward architectures that infer depth, camera pose, and 3D structure directly from images or videos. VGGT~\cite{vggt} introduced a unified backbone predicting dense geometry and camera parameters from unconstrained inputs. Its follow-ups further extend this paradigm: $\pi^3$~\cite{wang2025pi3} removes the reference-view bias with a permutation-equivariant design, ViPE~\cite{huang2025vipe} adapts geometry estimation to dynamic video scenes with large-scale annotated data, and MapAnything~\cite{keetha2025mapanything} unifies diverse 3D reconstruction tasks into a single transformer model. These advances demonstrate a clear trend toward scalable and consistent visual geometry learning.
\section{Methodology}
\label{sec-methodology}
Our objective is to develop a controllable video generation framework that enables precise manipulation of camera viewpoints in complex and dynamic scenes, without requiring explicit camera annotations. The design of our method is guided by two key principles:  
(a) \textbf{Consistency} — The model must leverage consistent camera poses and depth maps while maintaining coherence between the training and testing regimes;  
(b) \textbf{Extensibility} — the framework should readily extend to additional forms of control, such as inpainting or sketch guidance~\cite{vace}. Meanwhile, the system should be computationally efficient to train while preserving the expressive power of the underlying generative backbone. An overview of the proposed architecture is shown in~\Cref{fig-model-ours}. Our architecture is instantiated under the image-to-video (I2V) setting following \unic~\cite{cao2025uni3c}. 


\subsection{\modelname Tokenizer}
\label{sec-tokenizer}
\begin{figure*}[t!]
\begin{center}
\includegraphics[width=1.0\linewidth]{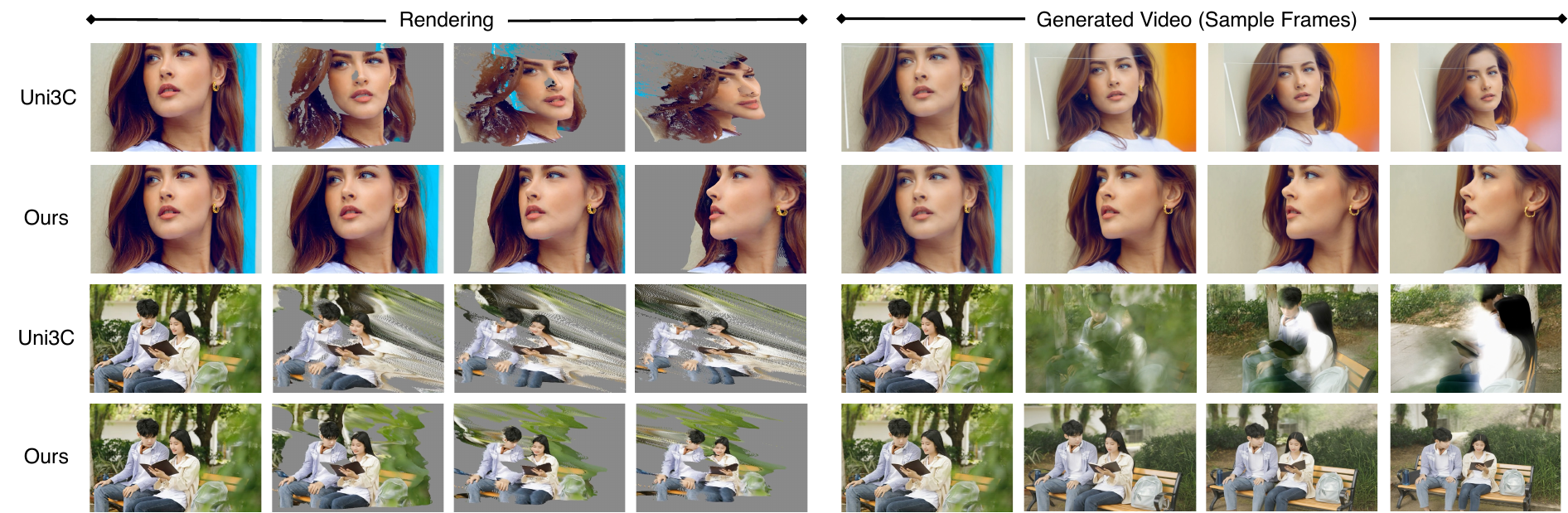}
\end{center}
\vspace{-10pt}
\caption{
\textbf{Comparison with the closest concurrent work \unic~\cite{cao2025uni3c}.} 
\textit{Left:} \unic renderings fail to accurately follow the intended camera motion and exhibit geometric distortions and outliers due to inconsistent 3D estimation. 
\textit{Right:} These inaccuracies in renderings lead to spatial misalignment and visible artifacts in the generated videos of \unic, while our generated videos do not exhibit these artifacts.
}
\label{fig-compare-uni3C}
\vspace{-6pt}
\end{figure*}

As illustrated in~\Cref{fig-model-ours}(a), the \modelname Tokenizer transforms raw, in-the-wild training videos (during training) or a reference frame with corresponding camera poses (during inference) into compact, geometry-aware representations that encode visual content and camera motion. 
\vspace{-4mm}
\paragraph{Training-time Tokenization.} Unlike prior controllable video generation baselines~\cite{cao2025uni3c} that require explicit camera annotations, we employ the Visual Geometry Grounded Transformer (\vggt)~\cite{vggt} to estimate both depth maps and camera parameters directly from raw videos. Given an input video sequence $\{I_t\}_{t=1}^T$ with $T$ frames, we first predict per-frame depth maps $\{\mathbf{D}_t\}_{t=1}^T$, corresponding camera extrinsics $\{\mathbf{E}_t\}_{t=1}^T$ and intrinsics $\{\mathbf{K}_t\}_{t=1}^T$ via a pre-trained frozen \vggt\cite{vggt} model:
\begin{equation}
\{\mathbf{D}_t, \mathbf{E}_t, \mathbf{K}_t\}_{t=1}^T = \mathrm{VGGT}^{\text{(frozen)}}\!\big(\{I_t\}_{t=1}^T\big).
\label{eq-vggt-train}
\end{equation}
This estimation ensures that the predicted depths and poses are geometrically aligned across frames and uniform across diverse training datasets, overcoming inconsistencies typically found in camera label annotations~\cite{vggt}. After that, a \textbf{point cloud reprojection} procedure is introduced to project the first frame's pixels into other frames, generating a rendering and a mask. We refer the readers to~\Cref{fig-compare-uni3C} for examples of renderings. Inferring both a rendering and a mask is central to many concurrent approaches, including inpainting-based video generation~\cite{cao2025uni3c,Ex4D} and generation conditioned on anchor videos~\cite{epic}. Specifically, the depth map $\mathbf{D}_{1}$ and camera intrinsics $\mathbf{K}_1$ of the first frame are used to back-project each pixel $\mathbf{y}^{(i)}_1 = [u^{(i)}, v^{(i)}, 1]^\top$ in the image plane of $I_1$ into 3D space, yielding a 3D point $\mathbf{X}^{(i)}$: 
\begin{equation}
\mathbf{X}^{(i)} = \mathbf{D}_1(\mathbf{y}^{(i)}_1)\, \mathbf{K}_1^{-1} \mathbf{y}^{(i)}_1, \quad \mathbf{X}^{(i)} \in \mathbb{R}^3.
\label{eq-backproject}
\end{equation}
We then map $\mathbf{X}^{(i)}$ into the coordinate frame of $I_t$ via the predicted extrinsics $\mathbf{E}_t = [\mathbf{R}_t \,|\, \mathbf{T}_t]$ and re-project it onto the image plane of $I_t$ using $\mathbf{K}_t$:
\begin{equation}
\tilde{\mathbf{y}}_t^{(i)} \sim \mathbf{K}_t \big(\mathbf{R}_t \mathbf{X}^{(i)} + \mathbf{T}_t \big).
\label{eq-reproject}
\end{equation}
The projected coordinates $\tilde{\mathbf{y}}_t^{(i)}$ define the image-plane sampling locations in frame $I_t$.
By assigning each $\tilde{\mathbf{y}}_t^{(i)}$ the color value from its corresponding 3D point $\mathbf{X}^{(i)}$ in $I_1$,
we obtain a synthesized view $\tilde{I}_t$—referred to as the \textbf{rendering}.
Meanwhile, we construct a binary visibility \textbf{mask} 
$\mathbf{M}_t \in \{0,1\}^{H \times W}$ for each frame based on point correspondences and occlusion checks, 
retaining only pixels whose projections $\tilde{\mathbf{y}}_t^{(i)}$ are traced from the reference frame $I_1$, where $H$ and $W$ are height and width, respectively:
\begin{equation}
\mathbf{M}_t(\tilde{\mathbf{y}}_t^{(i)}) = 
\begin{cases}
1, & \text{if } \tilde{\mathbf{y}}_t^{(i)} \text{ is visible and traced from } I_1,\\[2pt]
0, & \text{otherwise.}
\end{cases}
\label{eq-mask}
\end{equation}
The resulting mask $\mathbf{M}_t$ helps preserve regions reprojected from $I_1$ while filtering out occluded or unmatched areas.

The renderings $\tilde{I}$ and corresponding masks $\mathbf{M}$ are concatenated along the channel dimension and processed by a learnable \modelname Embedder $\mathcal{E}_{\Theta_{emb}}$. The \modelname Embedder follows the same architecture as the Wan video embedder~\cite{wan2025,vace}. We get the encoded rendering-mask as $\mathbf{z}_t^{(\mathrm{rm})} \in \mathbb{R}^{T\times H\times W \times d}$ where $d$ is the feature dimension:
\begin{equation}
\mathbf{z}_t^{(\text{rm})} 
= \mathcal{E}_{\Theta_{\text{emb}}}\left(\mathrm{Concat}_{\text{feat}}\big(\tilde{I}_t,\, \mathbf{M}_t\big)\right).
\label{eq:embedder}
\end{equation}
The embedding weights corresponding to $\tilde{I}_t$ are initialized from the Wan video embedder~\cite{wan2025,vace}, and the remaining weights corresponding to the mask $\mathbf{M}_t$ are zero-initialized.

Furthermore, we encode camera poses to augment the render-mask tokens $\mathbf{z}_t^{(\text{rm})}$. Following \vggt~\cite{vggt}, we represent the camera parameters using a quaternion-based concise form of the extrinsic and intrinsic matrices, denoted by $\mathbf{g}(\mathbf{E}_t, \mathbf{K}_t) = [\mathbf{q}_t, \mathbf{T}_t, \mathbf{f}_t]$, where $\mathbf{q}_t\in \mathbb{R}^4, \mathbf{T}_t\in \mathbb{R}^3, \mathbf{f}_t\in \mathbb{R}^2$ are the rotation quaternion, the translation vector, and the field of view, respectively. These vectors are concatenated in the feature dimension and passed through a linear layer to get a camera parameter embedding $\mathbf{z}_t^{(\mathrm{pr})} \in \mathbb{R}^{d}$, while $\mathbf{W}^{(\mathrm{pr})}$ and $\mathbf{b}^{(\mathrm{pr})}$ are learnable linear weights:
\begin{equation}
\mathbf{z}_t^{(\mathrm{pr})} 
= \mathbf{W}^{(\mathrm{pr})}\, \mathrm{Concat}\big(\mathbf{q}_t,\, \mathbf{T}_t,\, \mathbf{f}_t\big) 
+ \mathbf{b}^{(\mathrm{pr})}.
\label{eq-camera-embed}
\end{equation}

Finally, we concatenate $\mathbf{z}_t^{(\mathrm{pr})}$ with the visual tokens $\mathbf{z}_t^{(\mathrm{rm})}$ along the sequence dimension to form the final \textbf{\modelname token} $\mathbf{z}_t^{(\modelname)} \in \mathbb{R}^{(THW + 1) \times d}$ that jointly encodes visual content and camera geometry:
\begin{equation}
\mathbf{z}_t^{(\modelname)} = \mathrm{Concat}_\text{seq}\big(\mathbf{z}_t^{(\mathrm{rm})},\, \mathbf{z}_t^{(\mathrm{pr})}\big).
\label{eq-final-token}
\end{equation}

\vspace{-4mm}
\paragraph{Test-time Tokenization.} During inference, the model takes a reference frame $I_1$ as input, along with the camera parameters for all testing frames $\{\mathbf{E}_t, \mathbf{K}_t\}_{t=1}^{T}$. A single-view depth map $\mathbf{D}_1$ corresponding to the reference frame is estimated using \vggt~\cite{vggt}. A 3D point cloud is then derived using the back-projection formula (\Cref{eq-backproject}). After that, renderings, masks, and \modelname tokens are derived following~\Cref{eq-reproject,eq-mask,eq-camera-embed,eq-final-token}.
\begin{figure*}[t!]
\begin{center}
\includegraphics[width=1.0\linewidth]{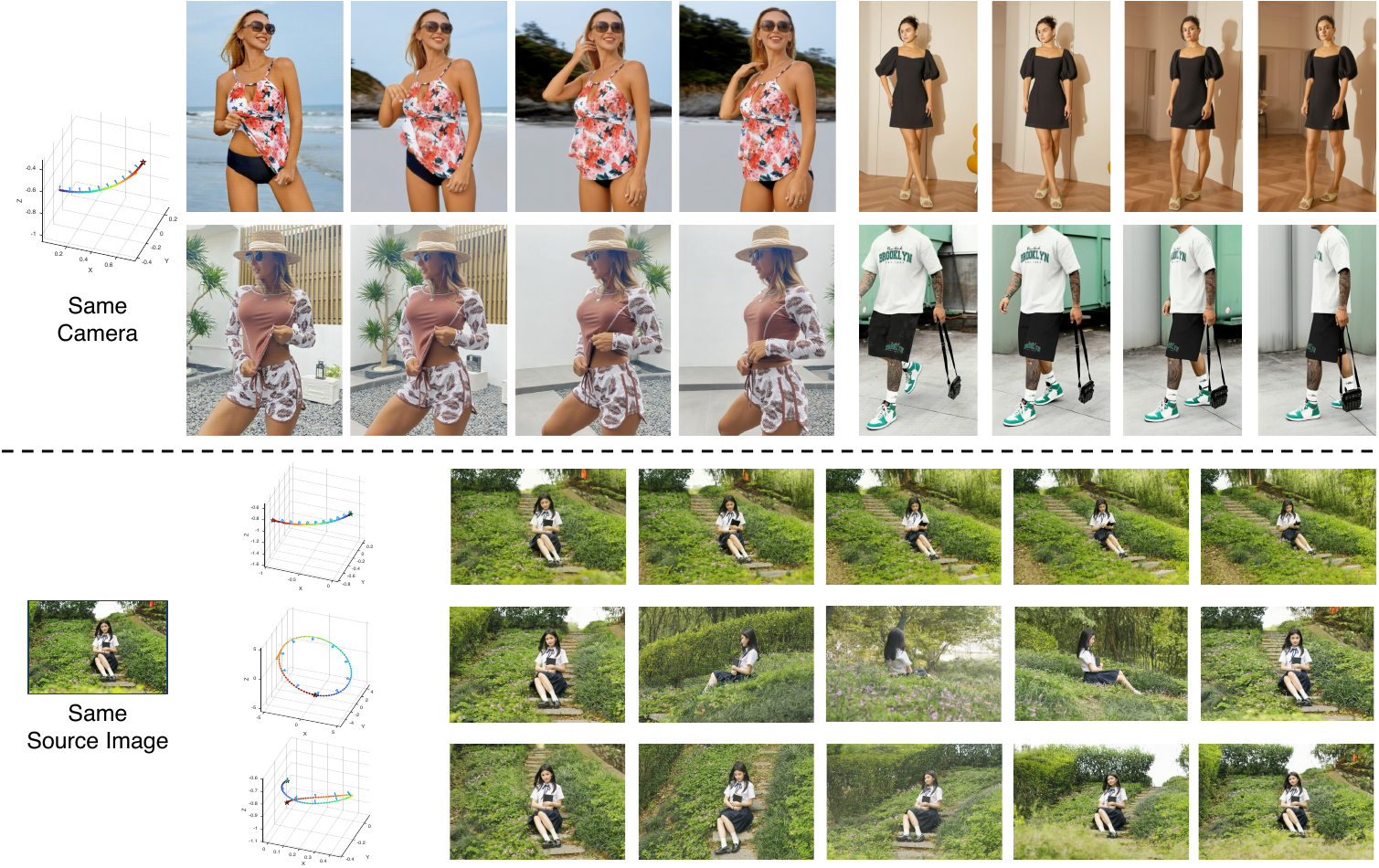}
\end{center}
\vspace{-20pt}
\caption{
\textbf{Camera control results.}
\textit{Top:} Four videos generated under the same camera trajectory (source images omitted for brevity).
\textit{Bottom:} Results with the same source image but different camera trajectories, illustrating consistent control across four distinct motions.
}
\label{fig-result-qualitative}
\vspace{-6pt}
\end{figure*}
\subsection{Token-Based Controlled Video Generation}
\label{sec-controlledgen}
As shown in~\Cref{fig-model-ours} (b), building upon the frozen pre-trained Wan DiT blocks, we introduce a token-based conditioning mechanism that integrates \modelname tokens with other control modalities in a unified latent space. In particular, our generation pipeline operates on four token types:  
(1) \textit{\modelname tokens} carrying camera poses and 3D geometry context.
(2) \textit{Noisy latents}, which are the initial Gaussian noises, are iteratively denoised during the diffusion process.
(3) \textit{Visual Condition Unit (VCU) tokens} represent the prompt, first frame, and mask. Here, the mask indicates all pixels to be modified by either the \modelname control or the \vace control. Noisy latents and VCU tokens already exist in the base Wan model.
(4) \textit{\vace tokens}~\cite{vace} encoding versatile external conditions such as layout or sketch. \vace tokens are not added if only the camera control is added.

These tokens are integrated into the base video diffusion model~\cite{wan2025}. Specifically, the \modelname tokens are processed by a set of learnable \textbf{\modelname Context Blocks}. Their outputs are projected through zero-initialized linear layers to align the feature dimensions with those of the Wan DiT blocks. The resulting embeddings are then added to the Wan DiT blocks~\cite{wan2025}, except for the token corresponding to the camera, which is excluded before the addition. In parallel, noisy latents and VCU tokens are injected into the Wan DiT blocks. Frozen \vace context blocks are also added to the Wan DiT blocks, conditioned jointly on the text embeddings and \vace tokens.

\subsection{Dataset Construction and Training Strategy}
\label{sec-training}
Training data plays a pivotal role in building effective video generative models~\cite{wan2025}. A key advantage of \modelname is that it can be trained directly on raw videos without requiring any camera pose annotations. We adopt a two-phase training strategy to progressively develop camera control capability while improving the generative quality.

In \textbf{Phase One}, we focus on learning general camera control from large-scale, diverse data. We collect a dataset of approximately 330K online videos and apply two filtering steps: (1) removing static-camera videos based on camera motion estimated by \vggt, and (2) discarding videos with low aesthetic quality measured by VBench~\cite{vbench}. After filtering, the resulting dataset contains roughly 100K videos. We also incorporate \vace-Benchmark to preserve the controlability of \vace tokens. Further details of this collection and filtering process are provided in~\Cref{appendix-sec-training-data-filter}.

In \textbf{Phase Two}, we emphasize fine-grained generation quality using a smaller, high-fidelity dataset. The training data consists of the training split of CameraBench~\cite{camerabench}, the I2V training set of \vace-Benchmark and our collected~\datasetname benchmark, totaling about 3K high-quality videos. The~\datasetname benchmark is generated from a state-of-the-art video foundation model Kling 2.5~\cite{klingai2024}, while images are our collected high quality human object interaction (HoI) images (see dataset construction details in~\Cref{appendix-sec-our-dataset}).

For both phases, the training parameters are the same (marked with a fire symbol in~\Cref{fig-model-ours}). Besides, the training targets are also the same, where we minimize the standard video flow matching loss~\cite{wan2025}:
\begin{equation}
\mathcal{L}_{\text{vel}} 
= \mathbb{E}_{t, \mathbf{x}_0, \mathbf{x}_1, c}
\Big[
\big\|
\mathbf{v}_\theta(\mathbf{x}_t, t, c) - (\mathbf{x}_1 - \mathbf{x}_0)
\big\|_2^2
\Big],
\label{eq:velocity-pred}
\end{equation}
where $\mathbf{x}_0 \!\sim\! \mathcal{N}(\mathbf{0}, \mathbf{I})$ is a Gaussian noise latent, 
$\mathbf{x}_1 \!=\! \mathcal{E}(I)$ is the encoded clean latent from the 3D VAE encoder, 
and $\mathbf{x}_t \!=\! (1-t)\mathbf{x}_0 + t\mathbf{x}_1$ interpolates between them at timestep $t\!\in\![0,1]$. 
The model $\mathbf{v}_\theta(\mathbf{x}_t, t, c)$ predicts the velocity field conditioned on $c$, 
which includes the text embedding and the input frame condition. Through this unified objective, \modelname\ progressively learns both visual fidelity and geometric consistency across training phases, which ensures stable optimization and seamless alignment between camera motion and scene dynamics.

\begin{figure*}[t!]
\begin{center}
\includegraphics[width=1.0\linewidth]{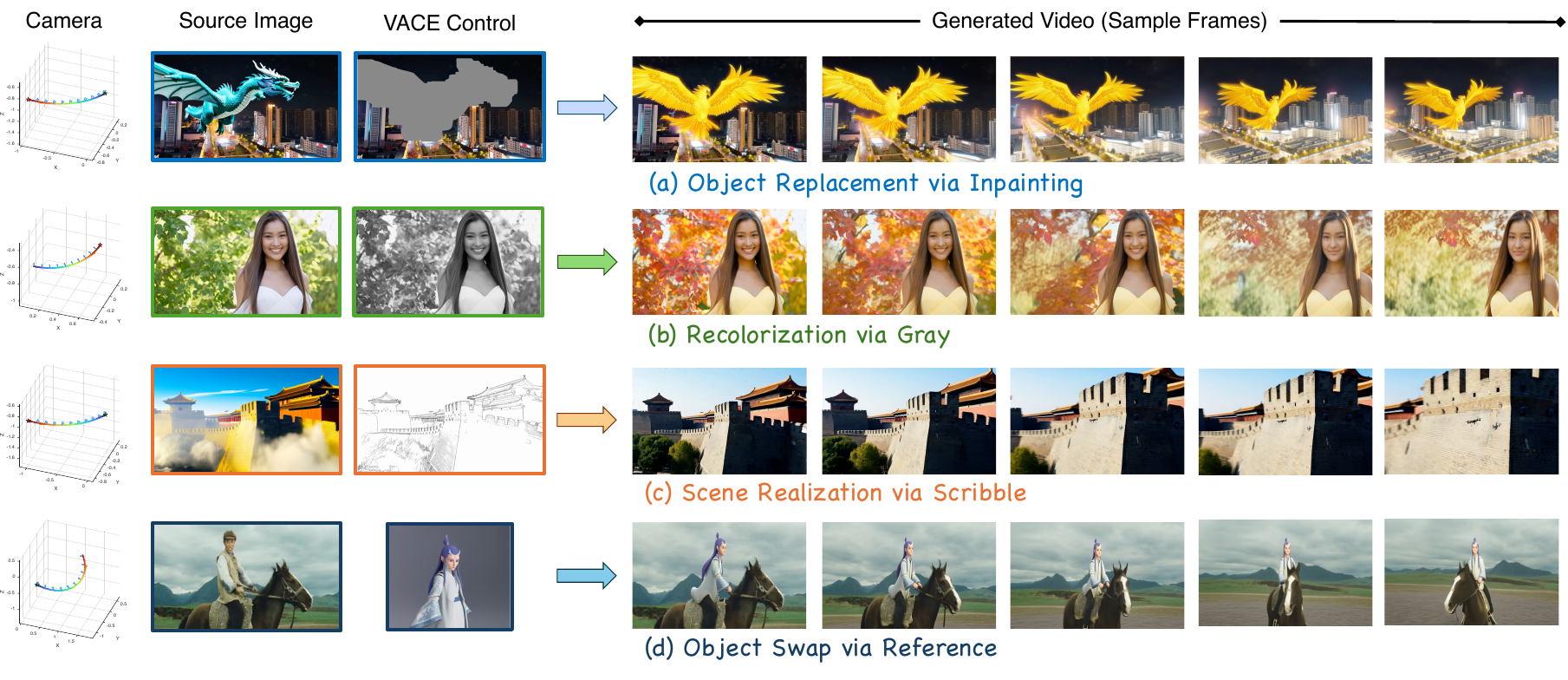}
\end{center}
\vspace{-20pt}
\caption{
\textbf{Extensibility Results.}
We show visualization results on controlling camera motion while achieving additional control through \vace~\cite{vace}. 
(a) \textit{Object replacement} (with prompt “replace the dragon with a phoenix”) using a source image and a masked image as \vace control, 
(b) \textit{Recolorization} of a green portrait guided by a gray image control, 
(c) \textit{Realization} of a virtual scene from a scribble as \vace control, and 
(d) \textit{Object swap} guided by a provided reference image. Please refer to the supplementary for more demo videos.}
\label{fig-result-vace}
\vspace{-6pt}
\end{figure*}
\begin{table*}[t!]
\centering
\resizebox{\textwidth}{!}{
\begin{tabular}{c|ccccccc|ccc|c}
\toprule
& Overall & Subject & Background & Aesthetic & Imaging & Temporal & Motion & ATE$\downarrow$ & RPE$\downarrow$ & RRE$\downarrow$ & Human Eval$\uparrow$ \\
& Score   & Consistency & Consistency & Quality & Quality & Flickering & Smoothness &  &  &  &  \\ \midrule
\multicolumn{12}{c}{\textbf{\unic-OOD-Challenging Dataset~\cite{cao2025uni3c}}} \\ \midrule
CameraCtrl~\cite{cameractrl} & 74.35 & 78.62 & 83.17 & 49.22 & 57.30 & 86.04 & 89.78 & 1.982 & 0.903 & 7.412 & 70.1 \\
SEVA~\cite{seva}             & 80.67 & 85.42 & 89.35 & 54.06 & 63.52 & 92.74 & 95.86 & 1.212 & 0.562 & 5.138 & 81.4 \\
EPiC~\cite{epic}             & 78.92 & 83.04 & 87.22 & 52.38 & 61.47 & 90.11 & 93.56 & 1.486 & 0.691 & 6.307 & 78.6 \\
\unic~\cite{cao2025uni3c}    & 82.75 & 88.19 & 91.03 & 57.61 & 66.90 & 95.40 & 97.39 & 1.003 & 0.425 & 4.452 & 84.9 \\
\rowcolor{gray}\modelname~(Ours) & \textbf{85.84} & \textbf{90.92} & \textbf{93.68} & \textbf{60.71} & \textbf{72.84} & \textbf{97.89} & \textbf{99.56} & \textbf{0.812} & \textbf{0.301} & \textbf{3.674} & \textbf{90.3} \\ \midrule

\multicolumn{12}{c}{\textbf{CameraBench Dataset~\cite{camerabench}}} \\ \midrule
CameraCtrl~\cite{cameractrl} & 75.64 & 80.33 & 84.52 & 49.85 & 59.04 & 87.93 & 91.21 & 1.864 & 0.872 & 7.109 & 72.4 \\
SEVA~\cite{seva}             & 81.23 & 86.19 & 89.66 & 54.71 & 64.52 & 93.51 & 96.22 & 1.184 & 0.534 & 5.024 & 82.0 \\
EPiC~\cite{epic}             & 79.42 & 84.57 & 88.12 & 53.08 & 62.48 & 91.02 & 94.13 & 1.402 & 0.648 & 6.018 & 79.8 \\
\unic~\cite{cao2025uni3c}    & 83.42 & 88.67 & 90.92 & 58.16 & 69.85 & 96.12 & 98.03 & 0.972 & 0.416 & 4.321 & 86.1 \\
\rowcolor{gray}\modelname~(Ours) & \textbf{86.12} & \textbf{91.84} & \textbf{93.54} & \textbf{61.04} & \textbf{73.12} & \textbf{97.96} & \textbf{99.48} & \textbf{0.793} & \textbf{0.286} & \textbf{3.622} & \textbf{91.5} \\ \midrule

\multicolumn{12}{c}{\textbf{\datasetname Benchmark (Ours)}} \\ \midrule
CameraCtrl~\cite{cameractrl} & 78.55 & 82.37 & 86.94 & 50.23 & 65.28 & 90.14 & 93.57 & 1.622 & 0.753 & 6.102 & 75.3 \\
SEVA~\cite{seva}             & 82.71 & 87.84 & 90.41 & 54.39 & 70.26 & 94.82 & 97.35 & 1.041 & 0.493 & 4.971 & 84.2 \\
EPiC~\cite{epic}             & 81.05 & 85.63 & 88.79 & 53.18 & 68.15 & 92.73 & 95.64 & 1.283 & 0.618 & 5.762 & 81.9 \\
\unic~\cite{cao2025uni3c}    & 84.36 & 90.03 & 91.39 & 55.03 & 73.50 & 97.00 & 99.19 & 0.941 & 0.392 & 4.206 & 87.8 \\
\rowcolor{gray}\modelname~(Ours) & \textbf{87.24} & \textbf{92.81} & \textbf{93.57} & \textbf{58.62} & \textbf{76.02} & \textbf{98.06} & \textbf{99.58} & \textbf{0.768} & \textbf{0.259} & \textbf{3.414} & \textbf{93.4} \\ 
\bottomrule
\end{tabular}
}
\caption{
\textbf{Camera-controlled generation results on three benchmarks.} Metrics include VBench~\cite{vbench} scores (overall, subject and background consistency, aesthetic and imaging quality, temporal flickering, motion smoothness), and 3D pose metrics (ATE, RPE, RRE). 
We additionally report a human evaluation score measuring perceived realism and temporal consistency. Refer to~\Cref{sec-quantitative-comparisons} for more details.
}
\label{tab:uni3c-comparison}
\vspace{-5pt}
\end{table*}

\section{Experiments}
\subsection{Experimental Setups}
\paragraph{Training and Inference Configurations.} We adopt 14B Wan-I2V Diffusers 2.1~\cite{wan2025} as our base model. \modelname context blocks contain $20$ external DiT layers, while the total number of trainable parameters is around 1B. We scale multi-resolution videos to a resolution of $587\times 587$ with a total of $77$ frames, at an FPS of $16$. Since the Wan model tokenizes videos into spatiotemporal patches before sequential processing, the total number of tokens (i.e., overall spatiotemporal resolution) primarily determines the model’s input size. For each training phase, the learning rate is linearly warmed up to $10^{-5}$ over the first $1$K steps and then kept constant. Phase~I training runs for $100$K steps, while Phase~II training runs for an additional $10$K steps. The entire training process takes approximately one week on $32$~H100 GPUs. Text prompts are generated with Tarsier2~\cite{tarsier2} using a simple descriptive query. Inference is performed with classifier-free guidance set to $\texttt{cfg} = 5.0$. The generated videos last for 5 seconds, for ours and all baselines. To save GPU memory, we adopt Fully Sharded Data Parallel (FSDP)~\cite{fsdp} and Sequence Parallelism (SP)~\cite{sp} on the base model (please refer to~\Cref{appendix-model-paralleization} for details).
\vspace{-15pt}
\paragraph{Baselines.}
We compare with several strong baselines spanning the spectrum of controllable video generation methods:
(1) CameraCtrl~\cite{cameractrl}, which conditions video diffusion models on camera poses through depth- or point-based re-projection into the diffusion backbone;
(2) SEVA~\cite{seva}, a novel view synthesis framework for static scenes that incorporates structural cues to enhance viewpoint consistency;
(3) EPiC~\cite{epic}, a concurrent method employing a 3D rendering–based conditioning strategy with lightweight re-projection modules to encode camera geometry during generation; and
(4) \unic~\cite{cao2025uni3c}, another concurrent large-scale approach that leverages 3D re-rendering with provided camera annotations for camera-controllable video generation.
\vspace{-8pt}
\paragraph{Datasets.} During comparison, each method generates videos conditioned on the same input image and camera trajectories for a fair comparison. We use the following three benchmarks.
(1) \unic-OOD-Challenging~\cite{cao2025uni3c}, designed for out-of-distribution evaluation, which contains source images and texts without paired videos. This dataset includes extreme camera trajectories unseen during training, highlighting each model’s generalization ability.
(2) CameraBench~\cite{camerabench}, which provides balanced, high-quality samples with moderate camera motion and well-calibrated trajectories, focusing on temporal stability and visual realism.
(3) Our constructed \datasetname\ benchmark, which emphasizes high-fidelity human–object interactions and indoor scenes. For all experiments on \datasetname, we use the eight predefined camera trajectories released by \unic~\cite{cao2025uni3c}; for each source image, eight corresponding videos are generated following these trajectories to assess viewpoint controllability and geometric consistency. We report results on the held-out validation split for both CameraBench~\cite{camerabench} and \datasetname. 
\begin{table}[]
\centering
\resizebox{\columnwidth}{!}{
\begin{tabular}{l|ccc|c}
\toprule
\textbf{Variant} & \textbf{Overall$\uparrow$} & \textbf{ATE$\downarrow$} & \textbf{RPE$\downarrow$} & \textbf{Remark} \\ 
\midrule
\textbf{Full \modelname} & \textbf{86.1} & \textbf{0.79} & \textbf{0.29} & All modules enabled \\ \midrule
\textit{w/ Text-Informed Camera Embedding} & 85.7 & 0.91 & 0.36 & Add text-input in camera encoding \\ 
\textit{Concat \modelname Context Block Features} & 83.9 & 0.95 & 0.39 & Replace “add” with concatenation \\ 
\textit{Plücker Ray Tokens} & 84.4 & 0.92 & 0.35 & Use Plücker ray as camera tokens \\ 
\textit{w/o Visibility Mask} & 83.2 & 1.05 & 0.45 & Don't use visibility mask \\ 
\textit{w/o Camera Embeddings} & 85.8 & 0.90 & 0.38 & Only use the rendering and mask \\ 
\textit{Inconsistent Train-Test Depth Estimation} & 84.6 & 0.90 & 0.35 & Follow \unic's depth estimation \\ 
\textit{Annotated Camera Poses} & 83.7 & 0.84 & 0.34 & Poses are not consistent with depths\\
\bottomrule
\end{tabular}
}
\vspace{-10pt}
\caption{
\textbf{Further Ablation Study on CameraBench~\cite{camerabench}.}  
Each variant disables or replaces a specific design choice in \modelname.
}
\label{tab:ablation}
\vspace{-10pt}
\end{table}
\vspace{-3pt}
\subsection{Camera Control Qualitative Results}
\label{sec-qualitative-results}
We visualize our camera control results in~\Cref{fig-result-qualitative}.
The top part shows four videos generated under the same camera trajectory given different source images, demonstrating consistent and geometry-aware viewpoint motion. The bottom part presents results from a single source image with three distinct camera trajectories, where \modelname\ accurately follows each prescribed path while preserving scene structure and visual realism. As further illustrated in~\Cref{fig-compare-uni3C}, our method produces geometrically consistent renderings that faithfully follow camera motion, addressing the distortion and misalignment issues in \unic~\cite{cao2025uni3c}.
These results highlight our model’s strong controllability and temporal coherence across diverse scenes and motion patterns. Besides, as illustrated in the teaser (\Cref{fig-teaser}), our framework generalizes across diverse domains and enables wide-range camera manipulation, including full 360° orbit. 

\subsection{Quantitative Comparisons}
\vspace{2pt}
\label{sec-quantitative-comparisons}
We perform quantitative comparisons between our method and the baselines using three sets of evaluation metrics:
(1) Visual quality metrics from VBench~\cite{vbench}, which assess overall visual fidelity, subject and background consistency, aesthetic quality, imaging sharpness, temporal flickering, and motion smoothness;
(2) 3D geometric metrics~\cite{cao2025uni3c}, including Absolute Trajectory Error (ATE), Relative Pose Error (RPE), and Relative Rotation Error (RRE), to evaluate camera pose accuracy and geometric consistency; and
(3) Human evaluation scores (0–10 initial scores rescaled to 0–100), which measure perceived realism and temporal coherence based on the averaged ratings of 30 participants. Each participant evaluated 100 generated videos per model, which typically required about one hour per model.

As shown in~\Cref{tab:uni3c-comparison}, \modelname\ achieves the best overall performance across all benchmarks, outperforming prior methods in both visual and geometric metrics. It demonstrates accurate camera controllability, improved temporal stability, and robust generalization across diverse video domains, with human evaluation further confirming its superior realism, coherence, and perceptual quality.

\subsection{Extensibility Results}
\label{sec-generalization}
We evaluate the extensibility of \modelname\ under the image-to-video (I2V) setting by integrating our camera tokens with additional conditioning modalities provided by \vace~\cite{vace}.
As shown in~\Cref{fig-result-vace}, \modelname\ supports four compositional controls such as object replacement, recolorization, scene realization, and object swap, achieving coherent generation under dynamic camera motion.
These results highlight that the proposed token-based design scales effectively and remains highly compatible with multi-modal conditioning, enabling seamless integration of camera control with appearance- or structure-level guidance.
Quantitative comparisons are in the supplementary (\Cref{appendix-vace-comparison}).

\subsection{Further Ablation Studies}
\label{sec-ablation-study}
We conduct detailed ablations on the validation split of CameraBench~\cite{camerabench} to ablate the full \modelname model, as summarized in~\Cref{tab:ablation}. 
Replacing additive feature fusion with concatenation or substituting our quaternion-based camera tokenization (following \vggt~\cite{vggt}) with Plücker ray tokens both degrade performance, confirming the benefit of our token and context block design. 
Removing visibility masks leads to lower visual quality and higher geometric errors, highlighting their importance in filtering occluded regions. Incorporating text-informed camera embeddings~\cite{vace} does not lead to further improvement, likely because text–camera alignment introduces additional semantic ambiguity and optimization complexity beyond the scope of pure geometric control. Finally, using annotated camera poses or inconsistent depth estimators~\cite{cao2025uni3c} results in reduced consistency between geometry and motion, validating our design choice of learning from geometry grounded estimations in a unified and consistent manner.

\vspace{-3pt}
\section{Conclusions}
We presented \modelname, a geometry-grounded framework for controllable video generation that enables precise camera manipulation without requiring annotated camera poses.
By leveraging depth and pose estimation from geometry foundation models, our method learns from raw dynamic videos and integrates seamlessly with pretrained diffusion backbones through token-based conditioning.
Extensive experiments demonstrate that \modelname achieves superior geometric consistency, controllability, and visual realism compared to recent baselines, while remaining extensible to additional control mechanisms such as layout or inpainting. We hope this work provides a new foundation for unified and geometry-aware controllable video generation.

%% file: sec/3_finalcopy.tex



%% file: sec/X_suppl.tex
\clearpage
\setcounter{page}{1}
\maketitlesupplementary
\section{Phase I Training Data Filtering Details}
\label{appendix-sec-training-data-filter}

This section provides the full details of the filtering pipeline used to construct the Phase~I training dataset. Our goal is to obtain a large, diverse collection of raw Internet videos that exhibit meaningful camera motion, sufficient aesthetic quality, and reliable geometry estimates from VGGT~\cite{vggt}. Starting from an initial pool of approximately 330K crawled videos, the final filtered dataset contains about 100K clips. We obtain the captions of the training data from \cite{yuan2025tarsier2} and use them as prompts during Phase I training. We describe each filtering stage below.

\subsection{Source Collection and Initial Preprocessing}
We gather publicly accessible Internet videos covering a broad range of content types, including indoor and outdoor natural scenes, human--object interactions, cinematic B-roll, landscape footage, and casual handheld recordings. All videos are used in full accordance with their respective licenses, which permit research activities such as model training. We further verify that the dataset excludes all offensive, violent, or otherwise inappropriate material. No identity-specific content is intentionally collected. Each raw video undergoes the following preprocessing steps:

\begin{itemize}
    \item \textbf{Frame extraction:} Frames are uniformly re-sampled at 16 FPS.
    \item \textbf{Resolution normalization:} The longer image side is resized to $720$ pixels while preserving aspect ratio.
    \item \textbf{Clip segmentation:} Videos longer than 45 seconds are divided into 8--12 second clips via shot-boundary detection using HSV-gradient and color-histogram thresholds.
    \item \textbf{Discard rules:} Clips with effective resolution below 360p, severe compression artifacts, or fewer than 81 usable frames are removed.
\end{itemize}

This produces roughly 330K candidate clips for further filtering.

\subsection{Camera Motion Detection with \vggt}
Phase~I training requires videos with visible camera motion rather than static-camera footage. For each clip, we apply \vggt~\cite{vggt} to estimate per-frame extrinsic parameters $\{\mathbf{E}_t\}_{t=1}^{T}$, where $\mathbf{E}_t = [\mathbf{R}_t | \mathbf{T}_t]$.

\paragraph{Translation and Rotation Magnitudes.}
We compute the frame-to-frame translation:
\begin{equation}
\Delta_T(t) = \|\mathbf{T}_{t+1} - \mathbf{T}_t\|_2,
\end{equation}
and the quaternion-based rotation:
\begin{equation}
\Delta_R(t) = 2\arccos\left( |\langle \mathbf{q}_{t+1}, \mathbf{q}_t \rangle|\right).
\end{equation}
We use the mean values $\mu_T$ and $\mu_R$ as indicators of camera motion.

\paragraph{Static-Camera Rejection.}
A clip is removed if the motion magnitudes satisfy:
\begin{equation}
\mu_T < 0.002 \quad \text{and} \quad \mu_R < 0.5^\circ.
\end{equation}
We additionally discard clips with discontinuous, unstable, or invalid pose predictions (e.g., quaternion flips, NaNs).  
After this step, approximately 160K clips remain.

\subsection{Aesthetic Quality Filtering}
To ensure the model does not learn low-quality photometric statistics, we evaluate each clip using the VBench~\cite{vbench} aesthetic quality predictor. For each frame:
\begin{equation}
s_{\text{aesthetic}} = \frac{1}{T} \sum_{t=1}^{T} f_{\text{VBench}}(I_t).
\end{equation}
We discard clips with $s_{\text{aesthetic}} < 0.20$. Heuristic rules remove additional low-quality clips exhibiting excessive compression, over/underexposure, or artificial slideshow-like pan-zoom effects.  
This step removes approximately 60K clips.

\subsection{Filtering Summary}

\begin{table}[h]
\centering
\begin{tabular}{p{0.45\columnwidth} p{0.22\columnwidth} p{0.20\columnwidth}}
\toprule
\textbf{Stage} & \textbf{Videos} & \textbf{Reduction} \\
\midrule
Raw crawled videos & 330{,}421 & -- \\
Motion filtering & 160{,}002 & $-52\%$ \\
Aesthetic filtering & 100{,}021 & $-38\%$ \\
\bottomrule
\end{tabular}
\caption{Phase~I training data filtering summary.}
\label{tab:phase1-filter-summary}
\end{table}
Table~\ref{tab:phase1-filter-summary} summarizes the filtering pipeline. The final Phase~I dataset thus offers stable and meaningful camera motion and high aesthetic quality, providing a strong foundation for learning robust and generalizable camera control.

\section{\datasetname Benchmark Collection}
\label{appendix-sec-our-dataset}

This section provides details of how we construct the \datasetname\ benchmark used in Phase~II training and all evaluations involving human--object interactions (HoI) and high-fidelity indoor scenes. The goal of \datasetname\ is to create a high-quality, diverse, and geometry-rich dataset that complements CameraBench~\cite{camerabench} and Uni3C-OOD-Challenging~\cite{cao2025uni3c}. All videos in \datasetname\ are generated using the state-of-the-art commercial video foundation model Kling~2.5~\cite{klingai2024}, paired with carefully curated source images. The content focuses primarily on shopping-related scenarios with high commercial value. No real identity-specific content is included. Below, we describe each component in detail.

\subsection{Source Image Collection}
\label{sec-source-image}
Our pipeline starts with constructing a diverse image corpus emphasizing human–object interactions (HoI) and indoor settings. All images are sourced from publicly accessible datasets and copyright-free photography repositories that permit research usage. To ensure strong geometric cues and compatibility with controllable video generation, we apply the following preprocessing and filtering steps:

\begin{itemize}
    \item \textbf{Resolution requirement:} Images must have a minimum longer-side resolution of 720~px.
    \item \textbf{Foreground quality:} The main subject (human, object, or both) must be clearly visible, not heavily occluded, and free of motion blur.
    \item \textbf{Scene type diversity:} Images are selected across a broad set of categories including kitchen, living room, office, workshop, retail stores, and studio environments.
    \item \textbf{Safety filtering:} Images containing identifiable private individuals, minors, copyrighted characters, or sensitive content are manually removed.
\end{itemize}

A total of \textbf{500} high-quality HoI-focused images are preserved after filtering.

\subsection{Prompt Generation}
\label{sec-prompt-generation-kling}
Before generating videos, we first construct a text prompt for each input image. Each prompt contains two components: a content description that captures the scene semantics, and a camera-movement instruction that specifies the intended motion. Both components are generated jointly using a Tarsier-2~\cite{yuan2025tarsier2}. We feed the image into Tarsier-2 and, through carefully designed system instructions, ask the model to produce a prompt that (1) accurately reflects the visual content—especially human–object interaction, (2) includes plausible rich motion movements, (3) includes explicit, large-scale camera-movement directives. This design encourages the model to learn strong and diverse camera motions in the resulting training videos. Here we provide some sample prompts used to generate videos:

\begin{itemize}
    \item \textbf{Sample Prompt 1.} Camera: wide orbit left with extreme dolly out.
    Outdoors, captured in a medium shot, a woman stands near the ocean, holding a phone with a stylish case in her hand. The background features soft waves and a cloudy sky, creating a serene coastal vibe. She gently raises the phone to eye level, showcasing the case's design while maintaining a natural and relaxed posture. The overall atmosphere is calm and breezy, complementing the coastal setting.

    \item \textbf{Sample Prompt 2.} Camera: wide orbit right with extreme dolly in. Outdoors, captured in a bright, sunny medium shot, a woman sits poolside holding a sunscreen spray bottle. She gently sprays the product onto her leg and spreads it evenly, highlighting the ease of application and smooth texture. The scene conveys a carefree, summery mood, enhanced by the sparkling pool in the background.

    \item \textbf{Sample Prompt 3.} Camera: extreme dolly out with sweeping pan left.
    Indoors, captured in a medium shot, a child wearing a bright yellow sweater sits at a table with a glass of chocolate drink in front of them. The child picks up the glass with both hands and takes a sip, emphasizing the drink's creamy texture. The vibrant blue background and cheerful setting reinforce a playful and inviting atmosphere.
\end{itemize}

\subsection{Video Generation Using Kling~2.5~\cite{klingai2024}}
For each source image and each prompt with camera, we generate a 5-second video using the Kling~2.5 video generation model. Kling~2.5 is chosen due to its strong scene coherence, high-resolution synthesis, and stable handling of indoor scenes.  Each video cost about five US dollars. The generation protocol is standardized to maintain reproducibility:

\begin{itemize}
    \item \textbf{Input:} See~\Cref{sec-source-image}.
    \item \textbf{Prompt:} See~\Cref{sec-prompt-generation-kling}.
    \item \textbf{Model Settings:} 16~FPS, 5-second duration, default Kling sampling schedule.
\end{itemize}

We generated a dataset of $500$ videos. The dataset is randomly split into two parts, where $100$ videos are left for evaluation, while the remaining videos are used for training.

\begin{table*}[t!]
\centering
\resizebox{\textwidth}{!}{
\begin{tabular}{c|ccccccc|ccc|cc}
\toprule
\textbf{Method} &
\textbf{Overall} &
\textbf{Subject} &
\textbf{Background} &
\textbf{Aesthetic} &
\textbf{Imaging} &
\textbf{Temporal} &
\textbf{Motion} &
\textbf{ATE$\downarrow$} &
\textbf{RPE$\downarrow$} &
\textbf{RRE$\downarrow$} &
\textbf{Human-Gen$\uparrow$} &
\textbf{Human-Cam$\uparrow$} \\
\midrule
\multicolumn{13}{c}{\textbf{Task: Inpainting}} \\ \midrule
VACE~\cite{vace}
 & 84.91 & 89.34 & 92.17 & \textbf{61.42} & 71.33 & \textbf{98.14} & 97.28
 & 3.912 & 1.942 & 14.21
 & 91.2 & 12.3 \\
\rowcolor{gray}\modelname~(Ours)
 & \textbf{86.73} & \textbf{91.26} & \textbf{93.51} & 59.31 & \textbf{73.28} & 97.93 & \textbf{99.41}
 & \textbf{0.812} & \textbf{0.301} & \textbf{3.674}
 & \textbf{92.8} & \textbf{89.6} \\ 
\midrule

\multicolumn{13}{c}{\textbf{Task: Gray-to-Color}} \\ \midrule
VACE~\cite{vace}
 & 85.77 & 90.54 & 92.83 & \textbf{60.74} & 70.08 & 97.56 & 96.95
 & 3.504 & 1.721 & 12.92
 & 92.0 & 10.7 \\
\rowcolor{gray}\modelname~(Ours)
 & \textbf{86.12} & \textbf{91.04} & \textbf{93.08} & 58.91 & \textbf{72.74} & \textbf{97.85} & \textbf{99.44}
 & \textbf{0.793} & \textbf{0.286} & \textbf{3.622}
 & \textbf{91.9} & \textbf{90.4} \\
\midrule

\multicolumn{13}{c}{\textbf{Task: Scribble-to-Video}} \\ \midrule
VACE~\cite{vace}
 & 84.62 & 89.03 & 91.04 & 57.64 & 69.08 & \textbf{97.61} & 96.22
 & 4.221 & 2.104 & 15.01
 & 90.7 & 11.8 \\
\rowcolor{gray}\modelname~(Ours)
 & \textbf{85.57} & \textbf{90.52} & \textbf{92.64} & \textbf{57.83} & \textbf{71.30} & 97.44 & \textbf{99.10}
 & \textbf{0.864} & \textbf{0.319} & \textbf{3.811}
 & \textbf{91.4} & \textbf{88.7} \\
\midrule

\multicolumn{13}{c}{\textbf{Task: Reference Image}} \\ \midrule
VACE~\cite{vace}
 & 86.92 & 91.71 & \textbf{94.33} & \textbf{61.12} & 71.21 & 97.86 & 97.44
 & 3.712 & 1.863 & 14.83
 & 93.1 & 13.9 \\
\rowcolor{gray}\modelname~(Ours)
 & \textbf{87.24} & \textbf{92.81} & 93.57 & 58.62 & \textbf{76.02} & \textbf{98.06} & \textbf{99.58}
 & \textbf{0.768} & \textbf{0.259} & \textbf{3.414}
 & \textbf{93.4} & \textbf{92.7} \\
\bottomrule
\end{tabular}
}
\caption{
\textbf{Extensibility comparison between VACE and \modelname across four compositional control tasks.}
VACE achieves competitive or even stronger VBench scores in a few dimensions (e.g., aesthetic quality, temporal stability, or background consistency). 
However, VACE completely fails at camera-following, resulting in extremely poor pose metrics and Human-Cam scores. 
\modelname maintains competitive generation quality while delivering accurate and smooth 3D-consistent camera control across all tasks.
}
\label{tab:vace-extensibility}
\end{table*}

\section{Quantitative Comparison against \vace}
\label{appendix-vace-comparison}

To evaluate the extensibility of \modelname\ beyond camera control, we conduct a detailed quantitative comparison against the unified controllable video generation framework \vace~\cite{vace}. We benchmark both methods across four representative controllable-generation tasks: inpainting, gray-to-color, scribble-to-video, and reference-image conditioning. Source videos, prompts and videos are from the~\vace benchmark. For each task, we report VBench scores (overall, consistency, aesthetic quality, imaging quality, temporal stability, and motion smoothness), 3D camera metrics (ATE/RPE/RRE), and two-part human evaluation: \textit{Human-Gen} for overall perceptual quality and \textit{Human-Cam} for perceived camera-following accuracy and smoothness. When grading \textit{Human-Cam}, human participants are given a description of the input camera trajectory. Other settings are same as specified before in~\Cref{sec-quantitative-comparisons}.

Table~\ref{tab:vace-extensibility} shows that \vace\ achieves competitive—and occasionally higher—VBench scores on appearance-centric metrics such as aesthetic quality, temporal stability, or background consistency. This reflects its strong underlying video diffusion backbone and its ability to perform high-quality appearance editing under various conditioning modalities. However, \vace\ fundamentally lacks camera-following capability. Since its architecture does not incorporate explicit geometry grounding or pose-dependent conditioning, \vace\ produces videos that visually resemble the controlled input but do not track the prescribed camera trajectory. This results in extremely poor pose accuracy across all tasks, with ATE, RPE, and RRE one to two orders of magnitude worse than \modelname. The Human-Cam evaluation confirms this failure: participants consistently rated \vace's camera adherence as very low (\(\sim\)10--14), often reporting ``static-camera'' or ``incorrect motion'' artifacts. In contrast, \modelname\ maintains strong and consistent performance across both appearance and geometry domains. Overall, \modelname\ offers the best of both worlds: competitive generation quality and reliable camera-following behavior. Meanwhile, \vace\ remains effective as a general-purpose controllable generator but is fundamentally unsuitable for camera-based control due to its absence of 3D-consistent mechanisms.

\section{Model Parallelization Details.} 
\label{appendix-model-paralleization}
To save GPU memory, we implement two essential strategies: Fully Sharded Data Parallel (FSDP)~\cite{fsdp} and Sequence Parallelism (SP)~\cite{sp}. These two techniques are applied on the base Wan DiT blocks, not on the~\modelname Context Blocks or \vace Context Blocks. FSDP~\cite{fsdp} partitions model parameters, gradients, and optimizer states across multiple GPUs, substantially reducing per-device memory usage while preserving scalability. We enable mixed precision with \texttt{bf16} computation to further reduce activation memory. SP~\cite{sp} further distributes long sequence computations among devices, allowing efficient handling of large spatiotemporal token sequences during training. We use a SP size of two during our training.